\documentclass[letterpaper, 10 pt, conference]{conf/ieeeconf}  % Comment this line out if you need a4paper

\IEEEoverridecommandlockouts                              % This command is only needed if 
\usepackage{cite}
\usepackage{amsmath,amssymb,amsfonts}
\usepackage{algorithmic}
\usepackage{graphicx}
\usepackage{textcomp}
\usepackage{xcolor}
\usepackage{bm}
\usepackage{gensymb}
\usepackage{balance}
\usepackage{makecell}

\usepackage{url}
\usepackage[linkcolor=black,citecolor=black,urlcolor=black,colorlinks=true]{hyperref}

\renewcommand{\vec}[1]{\bm{#1}}
\renewcommand{\matrix}[1]{\begin{bmatrix} #1 \end{bmatrix}}
\renewcommand{\frame}[1]{\mathcal{F}_{#1}}
\newcommand{\mat}[1]{\bm{#1}}
\newcommand{\nR}[1]{\mathbb{R}^{#1}}
\newcommand{\transpose}{^\top}
\newcommand{\atanTwo}[1]{{\rm atan2}\left( #1\right)}
\DeclareMathOperator*{\argmin}{arg\,min}

\title{\LARGE \bf
Tilt-Ropter: A Fully Actuated Hybrid Aerial-Terrestrial Vehicle with Tilt Rotors and Passive Wheels
}

\author{Ruoyu Wang$^{1,2}$, Xuchen Liu$^{1,3}$, Zongzhou Wu$^{1}$, Zixuan Guo$^{1}$, Wendi Ding$^{1}$, Ben M. Chen$^{1}$%
\thanks{This work was supported by the Research Grants Council of Hong Kong SAR under Grants 14217922, 14209623, and 14209424. (Corresponding author: Wendi Ding.)}%
\thanks{$^{1}$All authors are with the Department of Mechanical and Automation Engineering, The Chinese University of Hong Kong, Shatin, Hong Kong (e-mail: \{rywang, xcliu, zzwu, zxguo, wd.ding\}@link.cuhk.edu.hk, bmchen@cuhk.edu.hk).}%
\thanks{$^{2}$Ruoyu Wang is also with the Faculty of Engineering, The University of Hong Kong, Pokfulam, Hong Kong.}%
\thanks{$^{3}$Xuchen Liu is also with the Peng Cheng Laboratory, Shenzhen, Guangdong, China.}%
}

\begin{document}

\maketitle
\thispagestyle{empty}
\pagestyle{empty}

%%%%%%%%%%%%%%%%%%%%%%%%%%%%%%%%%%%%%%%%%%%%%%%%%%%%%%%%%%%%%%%%%%%%%%%%%%%%%%%%
\begin{abstract}

In this work, we present Tilt-Ropter, a fully actuated hybrid aerial-terrestrial vehicle (HATV) that integrates tilt rotors with passive wheels to enable efficient multi-modal locomotion. Unlike conventional underactuated HATVs, the fully actuated design of Tilt-Ropter allows decoupled force and torque control, improving maneuverability and ground locomotion efficiency.

A unified nonlinear model predictive controller (NMPC) is developed to track reference trajectories, enforce non-holonomic constraints, and accommodate contact effects across locomotion modes, while ensuring actuator feasibility through dedicated control allocation. To address complex wheel-ground dynamics, an external wrench estimator is incorporated to provide real-time interaction wrench estimates.

The system is validated through simulation and real-world experiments, including seamless air-ground transitions and trajectory tracking tasks. Experimental results demonstrate low tracking errors in both modes and reveal a 92.8\% reduction in power consumption during ground locomotion compared with flight, highlighting the platform's potential for long-duration missions where energy efficiency is critical.

\end{abstract}

%%%%%%%%%%%%%%%%%%%%%%%%%%%%%%%%%%%%%%%%%%%%%%%%%%%%%%%%%%%%%%%%%%%%%%%%%%%%%%%%
\section{Introduction}

In recent years, micro air vehicles (MAVs) have been widely applied in diverse real-world scenarios to improve working efficiency and reduce the risk of human operators. However, endurance remains a significant limitation, especially for applications that demand persistent operation, such as large-scale infrastructure inspection or long-duration surveillance. In such missions, MAVs often consume substantial energy when traveling between waypoints, and their limited battery capacity prevents continuous task execution. To alleviate this issue, researchers have proposed HATVs \cite{kalantari2014modeling, zhang2022autonomous, fan2019autonomous, zhang2023model, wu2023unified, pan2023skywalker, cao2024aircrab, yu2025hybrid, yang2022sytab, qin2020hybrid, lin2024skater}, which combine the long-endurance advantage of ground vehicles with the mobility of aerial platforms. Most of these designs are based on quadrotors, leveraging their compact size and vertical takeoff and landing capability for seamless locomotion mode switching.

Despite these advantages, quadrotor-based HATVs suffer from inherent underactuation: attitude and position control cannot be decoupled. As a result, the vehicle must tilt its body to generate forward thrust, which produces undesired vertical forces during ground locomotion and reduces efficiency. To overcome such limitations of conventional quadrotor-based aerial robots, researchers have explored fully actuated MAVs (FAMAVs), which can generate arbitrary forces and torques in six degrees of freedom. This capability enables independent control of translation and rotation, making FAMAVs a promising basis for HATVs, particularly in scenarios requiring efficient ground locomotion or physical interaction \cite{hamandi2021design, bodie2020active, bodie2020towards}. Nevertheless, actuation redundancy introduces challenges in control allocation, increases mechanical complexity, and may reduce energy efficiency due to additional actuators and structures. Moreover, the coexistence of multiple locomotion modes further complicates the control problem.

\begin{figure}[!t]
    \vspace{6pt}
    \centering
    \includegraphics[width=1.0 \linewidth]{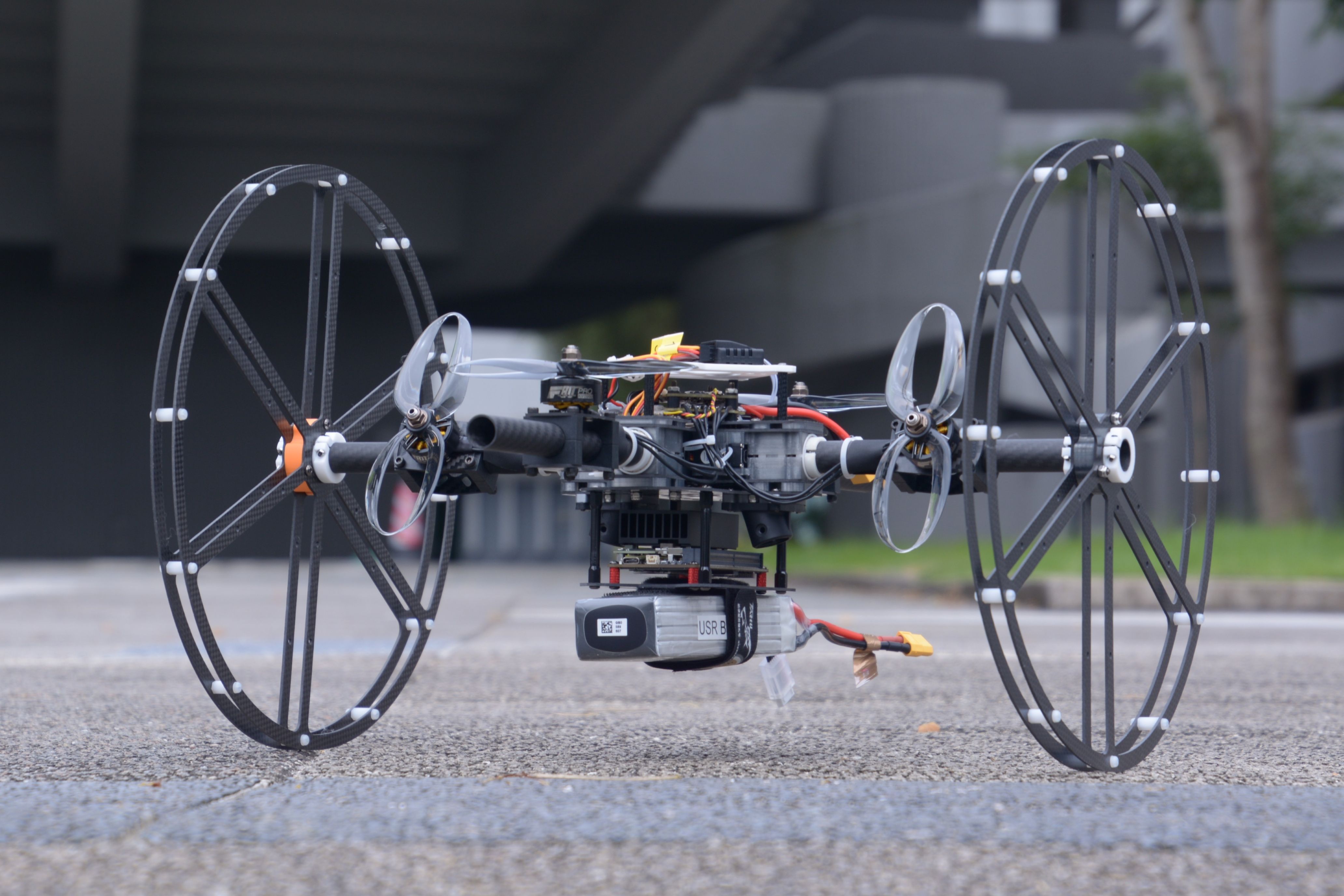}
    \caption{Tilt-Ropter: a fully actuated hybrid aerial-terrestrial vehicle with tilt rotors and passive wheels.}
    \label{fig:tilt_ropter}
    \vspace{-12pt}
\end{figure}

In this work, we present the design of \emph{Tilt-Ropter}, a fully actuated hybrid aerial-terrestrial vehicle that integrates tilt rotors with two passive wheels, as shown in Fig.~\ref{fig:tilt_ropter}. This configuration enables efficient locomotion both in the air and on the ground, with thrust generated directly along the direction of motion to improve maneuverability. To address the challenges of redundant actuation and multi-modal locomotion, we design a unified wrench-based nonlinear model predictive controller (NMPC) framework that incorporates actuator constraints and ground-specific non-holonomic conditions. To account for complex and difficult-to-model wheel-ground interaction forces and torques during terrestrial locomotion, we adopt an external wrench estimation framework that treats contact effects as disturbances rather than explicitly modeling friction and rolling dynamics. The estimated external wrench is incorporated into the dynamic model to improve tracking robustness under modeling uncertainties. The main contributions of this work are summarized as follows:
\begin{itemize}
    \item The design and implementation of a fully actuated hybrid aerial-terrestrial vehicle that integrates tilt rotors with passive wheels, enabling high maneuverability and energy-efficient locomotion.

    \item A unified wrench-based NMPC framework that seamlessly handles aerial and terrestrial locomotion, while respecting actuator limitations through a dedicated control allocation.

    \item An actuator-aware external wrench estimation method tailored to tilt-rotor aerial robots, enabling interaction wrenches and unmodeled dynamics to be estimated and incorporated into the model-based control framework.

\end{itemize}

%%%%%%%%%%%%%%%%%%%%%%%%%%%%%%%%%%%%%%%%%%%%%%%%%%%%%%%%%%%%%%%%%%%%%%%%%%%%%%%%
% Introduce related works
\section{Related Works}
\label{Related Works}
\subsection{Hybrid Aerial-Terrestrial Vehicles}

These platforms integrate aerial agility with ground mobility, allowing for energy-efficient long-distance traversal and agile obstacle avoidance via flight. Existing designs can be broadly classified by their ground locomotion mechanisms.

\subsubsection{Passive Wheel Designs}

Many HATVs adopt passive wheel structures to achieve terrestrial mobility with minimal mechanical complexity. Kalantari~\textit{et al.}~\cite{kalantari2014modeling} enclosed a quadrotor in a freely rotating cage, enabling ground motion without additional actuators. However, friction between the cage and ground reduces maneuverability. More recent works~\cite{fan2019autonomous, zhang2022autonomous, zhang2023model} introduced two side-mounted passive wheels, improving flexibility while keeping the additional mass low. Nonetheless, these designs remain underactuated and rely solely on differential thrust for yaw control, limiting precise ground steering.

To further reduce mechanical complexity, some designs employ a single passive wheel mounted on a quadrotor or bi-copter platform~\cite{pan2023skywalker, qin2020hybrid, yu2025hybrid}. Although this setup minimizes weight, it introduces static instability during terrestrial locomotion and often requires extra energy to maintain balance.

\subsubsection{Active Wheel Designs}

Active wheel mechanisms have been explored to enhance ground agility. Sihite~\textit{et al.} and Mandralis~\textit{et al.}~\cite{sihite2023multi, mandralis2025atmo} proposed systems with belt-driven wheels and differential steering, enabling omnidirectional motion and smooth transitions. However, the added mechanisms increase the system weight to over 5~kg, significantly compromising flight performance and endurance. Shi~\textit{et al.}~\cite{shi2024mtabot} introduced transformable wheels for flying, rolling, and climbing, with shared actuators to reduce weight, but the use of six actuators and timing belts adds design complexity and weight. A decoupled design was presented by Cao~\textit{et al.}~\cite{cao2023doublebee}, which combines a bi-copter and a two-wheeled differential drive. While offering independent control in both modes, its complexity and added mass limit flight agility.

\subsection{Fully Actuated Micro Air Vehicles}

Most HATVs are built upon underactuated quadrotors or bi-copters~\cite{cao2024aircrab, fan2019autonomous, zhang2023model, wu2023unified, zhang2022autonomous, pan2023skywalker, kalantari2014modeling, yu2025hybrid, yang2022sytab, qin2020hybrid, lin2024skater}, where the coupling of attitude and position control limits maneuverability, particularly in yaw and lateral motion. While bi-copters offer improved yaw control via rotor tilting~\cite{lin2024skater, yang2022sytab}, they still face constraints in agile multi-directional maneuvers.

While most HATVs remain underactuated, recent progress in the MAV field has produced fully actuated designs~\cite{ryll2014novel, ryll2016modeling, kamel2018voliro, li2024servo}, introducing actuation schemes beyond conventional quadrotors, with potential benefits for future HATV development. Such platforms can in principle generate arbitrary forces and torques~\cite{bodie2020active, bodie2020towards}, enabling precise 6-DoF control and improved interaction with the environment.

However, fully actuated designs introduce non-negligible trade-offs. The inclusion of additional actuators, such as tilting mechanisms, increases structural complexity, mass, and power consumption, which may compromise flight efficiency. Moreover, actuation redundancy requires dedicated control allocation strategies to resolve input redundancy.

\subsection{Control for Hybrid Aerial-Terrestrial Vehicles}

Controlling HATVs requires handling mode transitions and distinct dynamics in each mode. Model-free approaches, particularly PID controllers, are widely adopted for their simplicity~\cite{yu2025hybrid, guo2024multimodal, dong2024tactv, qin2020hybrid, pan2023skywalker, yang2022sytab, fan2019autonomous, zhang2022autonomous}, typically using a cascaded structure for position, attitude, and rate control. However, these methods often oversimplify ground interaction and lack robustness during mode transitions.

Model-based methods offer improved performance by leveraging accurate system dynamics~\cite{mandralis2025atmo, lin2024skater, wu2023unified, zhang2023model}. For example, Zhang~\textit{et al.}~\cite{zhang2023model} achieved trajectory tracking at up to 3~m/s, while Lin~\textit{et al.}~\cite{lin2024skater} demonstrated stable control on slippery terrain. Despite their advantages, most existing model-based methods are developed for underactuated platforms, whose control design is typically tailored to platform-specific actuation limits and the inherent coupling between translational and rotational motion. Therefore, these methods cannot be directly transferred to fully actuated HATVs, which require explicitly coordinating 6-DoF wrench generation, actuator allocation, and ground-contact constraints across locomotion modes. In this work, we aim to bridge this gap by presenting a fully actuated hybrid aerial-terrestrial platform along with a unified model-based control approach.

%%%%%%%%%%%%%%%%%%%%%%%%%%%%%%%%%%%%%%%%%%%%%%%%%%%%%%%%%%%%%%%%%%%%%%%%%%%%%%%%

\section{System Design}

\subsection{Mechanical Design}

Tilt-Ropter is a tilt-rotor quadcopter integrated with two passive wheels, enabling seamless transitions between aerial and terrestrial locomotion. The quadcopter comprises a central base frame and four articulated tilt arms. Each arm is equipped with a T-Motor F80 Pro brushless DC (BLDC) motor, paired with a Foxeer 5145 Donut toroidal propeller.

\begin{figure}[tbp]
    \vspace{6pt}
    \centering
    \includegraphics[width=0.95 \linewidth]{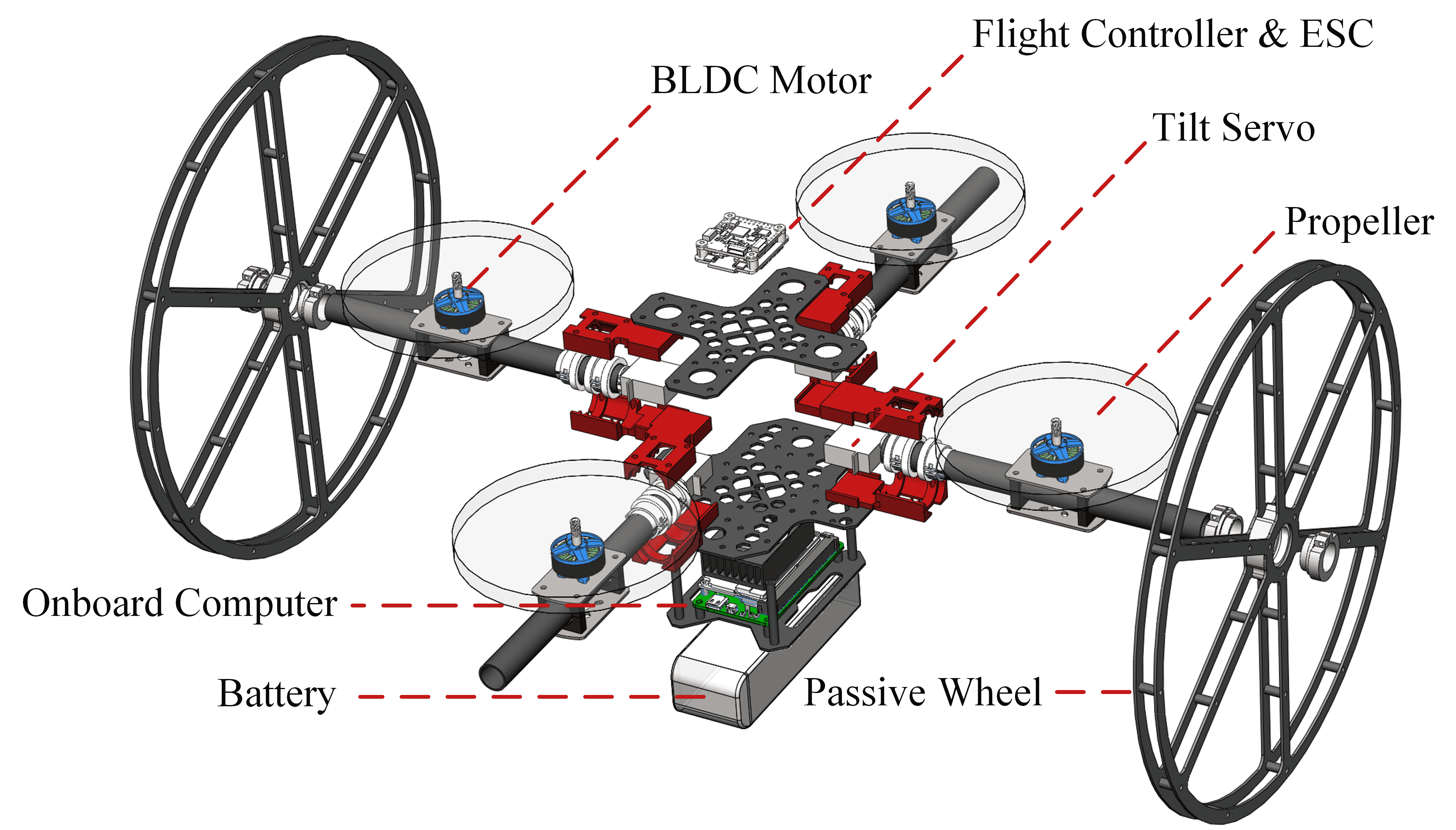}
    \caption{Exploded view of Tilt-Ropter key components.}
    \label{fig:tilt_ropter_exploded}
    \vspace{-12pt}
\end{figure}

The tilt mechanism for each arm is actuated by a GDW RS0708 servo motor, which is mounted within the base frame of the vehicle. Unlike conventional tilt-rotor MAV designs \cite{ryll2014novel, li2024servo}, where servo motors are typically positioned at the end of the tilt arms, Tilt-Ropter's configuration concentrates the mass closer to the center. This centralized layout minimizes the moment of inertia, enhancing agility and responsiveness. The detailed design of the robot is shown in Fig.~\ref{fig:tilt_ropter_exploded}. 

\begin{figure}[htbp]
    \vspace{-6pt}
    \centering
    \includegraphics[width=0.9 \linewidth]{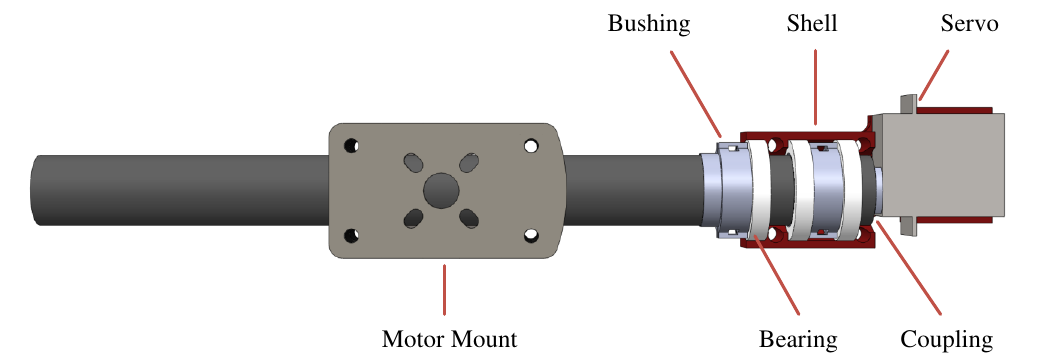}
    \caption{Illustration of the design of the tilt arm.}
    \label{fig:tilt_arm}
    \vspace{-6pt}
\end{figure}

While the centralized tilt configuration improves agility by reducing the moment of inertia, the tilting torque must be transmitted through the arm, which may introduce vibration due to structural clearance \cite{liu2023tj}. To mitigate this issue, ceramic bearings and a rigid 3D-printed support structure are adopted to eliminate backlash and enhance structural stiffness. This design ensures stable torque transmission and improves control performance during aggressive maneuvers, as illustrated in Fig.~\ref{fig:tilt_arm}.

Distinct from conventional quadrotor-based HATVs, which usually require additional supporting structures to mount wheels, Tilt-Ropter adopts an innovative design in which two opposing tilt arms serve directly as axles for the passive wheels. Since only the wheels themselves are added, this configuration greatly reduces both structural weight and overall complexity. Passive wheels were chosen for their simplicity, lightweight nature, and minimal additional energy consumption during multi-modal locomotion.

\begin{table}[htbp]
    \centering
    \caption{Models and Weights of Key Hardware Components}
    \begin{tabular}{l|l|l}
        \hline
        Component         & Model                        & Weight (g)  \\ 
        \hline
        Battery           & ACE 5300 mAh 45C             & 485         \\
        Servo motors      & GDW RS0708                   & 145         \\
        Onboard computer  & NVIDIA Jetson Orin NX        & 150         \\
        Brushless motors  & T-Motor F80 Pro              & 146         \\
        Propellers        & Foxeer 5145 Donut            & 20          \\
        ESC               & Holybro Tekko32 60A          & 14        \\
        Flight controller & Holybro Kakute H7 V1.3       & 8           \\
        \hline
    \end{tabular}
    \label{table:model_weight}
\end{table}

\subsection{Hardware System Design}

Tilt-Ropter employs a two-layer hardware architecture consisting of a low-level flight controller and a high-level onboard computer. The flight controller (Holybro Kakute H7 V1.3) acquires sensor data in real time and performs control allocation to convert high-level force and torque commands into motor and servo inputs. High-level tasks, including model-based control, state estimation, perception, and motion planning, are carried out on an NVIDIA Jetson Orin NX running Ubuntu 20.04 with ROS.

Table \ref{table:model_weight} details the models and weights of the key hardware components employed in the Tilt-Ropter. The fully assembled Tilt-Ropter, equipped with an ACE 5300 mAh 45C battery, weighs only about 1.5 kg, which is comparable to a conventional quadrotor.

%%%%%%%%%%%%%%%%%%%%%%%%%%%%%%%%%%%%%%%%%%%%%%%%%%%%%%%%%%%%%%%%%%%%%%%%%%%%%%%%

\section{Modeling}

\subsection{Frame Definition}

\begin{figure}[htbp]
    % \vspace{-6pt}
    \centering
    \includegraphics[width=0.9 \linewidth]{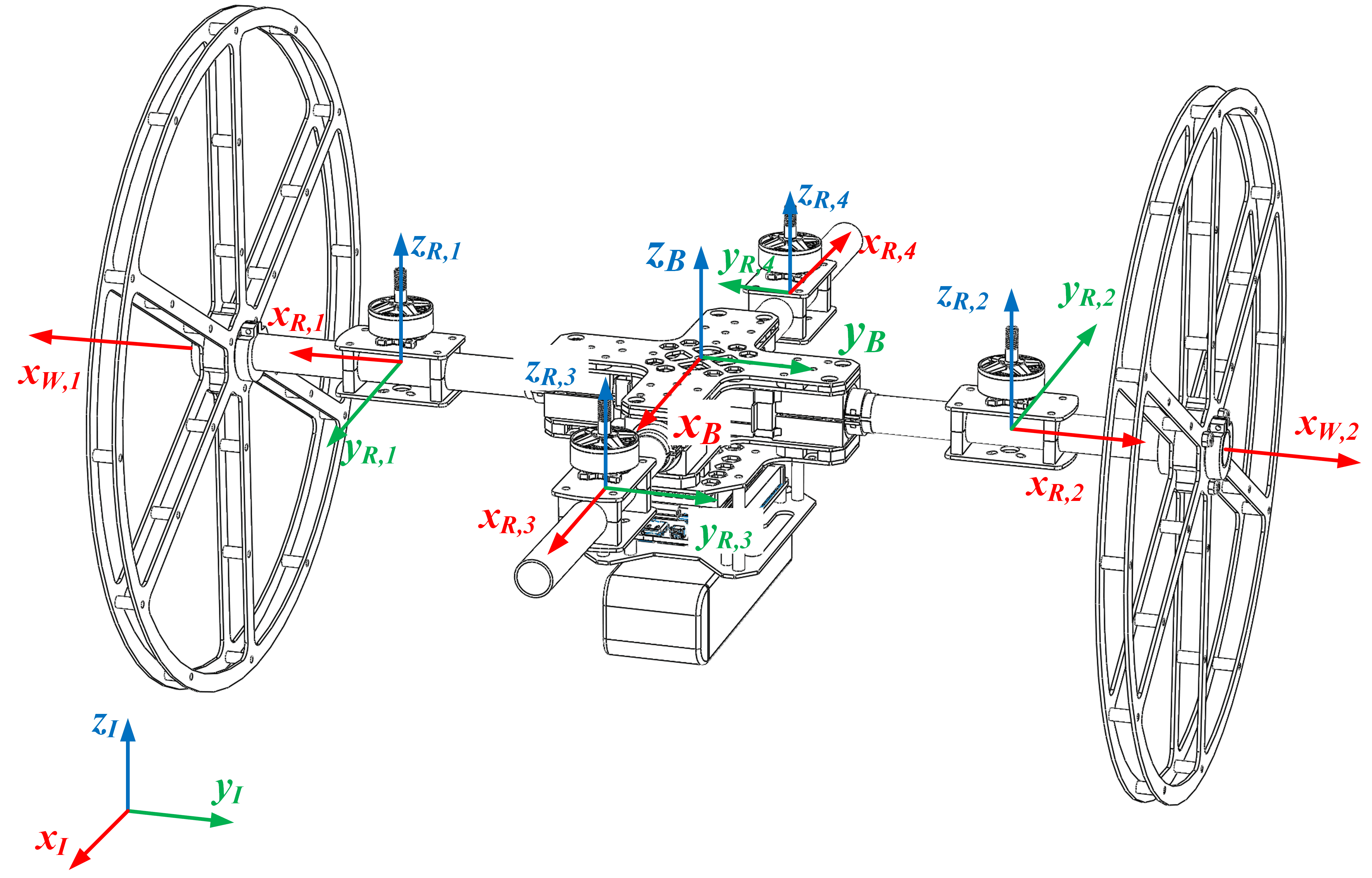}
    \caption{Definition of coordinate frames.}
    \label{fig:frame_all}
    \vspace{-6pt}
\end{figure}

Throughout this work, we introduce eight coordinate frames, as shown in Fig.~\ref{fig:frame_all}. The fixed East-North-UP (ENU) inertial frame is denoted by $\frame{I}$. The body frame, $\frame{B}$, is attached to the center of mass (CoM) of the robot. Four rotor frames, $\frame{R,i} \ (i = 1,2,3,4)$ are each defined with their origins at the intersection of the propeller's axis of rotation and the central axis of the respective arm. Rotor 1 and rotor 2 rotate in the counterclockwise direction, whereas rotor 3 and rotor 4 rotate in the clockwise direction. The $x$-axis of each $\frame{R,i}$ aligns with the central axis of its arm and points outward from the CoM. Additionally, two wheel frames, $\frame{W,i} \ (i = 1,2)$, are introduced, where each frame's origin is located at the geometric center of the wheel. The $x$-axis of each $\frame{W,i}$ is aligned with the $y$-axis of $\frame{B}$.

\subsection{Rigid Body Dynamics}

We assume that Tilt-Ropter can be modeled as a rigid body with mass $m$ and inertia $\bm{J}$, which is driven by the total force $\vec{F} \in \nR{3}$ and torque $\vec{M} \in \nR{3}$ of actuators acting on the CoM, and the gravity $\vec{g} = [0, 0, -9.81]^\top\,\mathrm{m\cdot s^{-2}}$.
\begin{subequations}
\begin{align}
{\dot{\vec{p}}} &= {\vec{v}}, \label{eq:rb1} \\
{\dot{\vec{q}}} &= {\frac{1}{2} \, \vec{q} \otimes \begin{bmatrix}
                                                0 \\ \vec{\omega}
                                     \end{bmatrix}}, \label{eq:rb2} \\
{\dot{\vec{v}}} &= {m^{-1} \vec{q} \odot  \left( \vec{F} + \vec{F}_e \right) + \vec{g}}, \label{eq:rb3} \\
{\dot{\vec{\omega}}} &= {\bm{J}^{-1}(\vec{M} + \vec{M}_e -\vec{\omega} \times (\bm{J} \vec{\omega}))}, \label{eq:rb4}
\end{align}
\end{subequations}
where $\otimes$ denotes quaternion multiplication and $\odot$ denotes the multiplication of a quaternion and a vector. $\vec{p} \in \nR{3}$ and $\vec{v} \in \nR{3}$ are the position and velocity of CoM in $\frame{I}$. $\vec{q} \in \nR{4}$ and $\vec{\omega} \in \nR{3}$ are the quaternion and angular velocity of the body frame $\frame{B}$.

The external force $\vec{F}_e \in \nR{3}$ and torque $\vec{M}_e \in \nR{3}$ are expressed in the body frame $\frame{B}$ and represent lumped disturbances. These disturbances include wheel-ground contact wrenches, friction effects, rolling resistance, and other environmental interactions that are not explicitly modeled in the rigid-body dynamics. Instead of constructing a detailed contact model, these effects are estimated online and incorporated into the model-based control framework for disturbance compensation.

\subsection{Control Allocation}
\label{sec:control allocation}

Tilt-Ropter is a fully actuated aerial vehicle capable of generating arbitrary six-dimensional wrenches. Unlike a conventional quadrotor with a fixed allocation matrix, the mapping from wrench to actuator commands in this platform is nontrivial. We define the total wrench generated by all actuators and applied at CoM as $\vec{W} = \begin{bmatrix} \vec{F}\transpose \vec{M}\transpose \end{bmatrix}\transpose \in \nR{6}$, which results from the combined effects of the rotor thrusts $T_i$ and the corresponding servo tilt angles $\alpha_i$ of all tilt-rotor units. A dedicated control allocation formulation is therefore required to map the desired wrench to feasible actuator inputs. This formulation also exposes implicit actuator constraints, such as limited servo rate, which will later be incorporated into the NMPC.

Here we neglect the angular acceleration and gyroscopic effects of the rotors, and the thrust $T_i$ is proportional to the square of the rotor speed $\Omega_i$. The relationship between the wrench command and actuator inputs is given by
\begin{equation}
    \begin{aligned}
        \vec{W} =
        &\begin{bmatrix}
            \vec{F} \\ \vec{M}
        \end{bmatrix}
        = \mat{A}  \vec{T}, \\
        \vec{T} =
        &\begin{bmatrix}
            T_{1,l} \\
            T_{1,v} \\
            \vdots \\
            T_{4,l} \\
            T_{4,v}        
        \end{bmatrix}
        = c_t
        \begin{bmatrix}
            \sin(\alpha_1) \Omega_1^2 \\
            \cos(\alpha_1) \Omega_1^2 \\
            \vdots \\
            \sin(\alpha_4) \Omega_4^2 \\
            \cos(\alpha_4) \Omega_4^2        
        \end{bmatrix},
    \end{aligned}
    \label{eq: allocation}
\end{equation}
where $\mat{A} \in \nR{6\times8}$ is the control allocation matrix, and $\vec{T}$ is an intermediate thrust vector defined in the body frame $\frame{B}$. The vector $\vec{T}$ represents the lateral and vertical components of each rotor thrust, thereby incorporating the effect of the servo angle $\alpha$, while $c_t$ denotes the thrust coefficient. This formulation simplifies $\mat{A}$ into a static matrix that depends only on the fixed geometry of the robot. Here, the small CoM and inertia variations induced by servo rotation are neglected, since the tilting units are lightweight relative to the vehicle and remain close to the nominal rotor locations. Consequently, $\vec{T}$ can be directly calculated using the Moore--Penrose pseudoinverse as
\begin{equation}
    \begin{aligned}
        \vec{T} &= \mat{A}^{\dagger} \vec{W}.
    \end{aligned}
    \label{eq: allocation_inverse}
\end{equation}
The thrust $T_i$ of the $i$-th rotor is calculated as
\begin{equation}
    \begin{aligned}
        T_i = \sqrt{T_{i,l}^2 + T_{i,v}^2}.
    \end{aligned}
    \label{eq: thrust_i}
\end{equation}
Then the rotor angular speed $\Omega_i$ and servo angle $\alpha_i$ can be obtained by:
\begin{equation}
    \begin{aligned}
        \Omega_i &= \sqrt{\frac{T_{i}}{c_t} }, \\
        \alpha_i &= \atanTwo{T_{i,l}, T_{i,v}}.
    \end{aligned}
    \label{eq: actuator_command}
\end{equation}
Given the relationship between the servo angle $\alpha$ and the thrust vector $\vec{T}$, the corresponding relationship between the servo angular velocity $\dot{\alpha}$ and the wrench change rate $\dot{\vec{W}}$ can be derived as
\begin{equation}
    \begin{aligned}
        \dot{\alpha}_i &= \frac{\dot{T}_{i,l} T_{i,v} - T_{i,l} \dot{T}_{i,v}}{T_{i,v}^2 + T_{i,l}^2}, \\
        \dot{T}_{i,l} &= \big(\mat{A}^{\dagger} \dot{\vec{W}}\big)_{2i-1}, \\
        \dot{T}_{i,v} &= \big(\mat{A}^{\dagger} \dot{\vec{W}}\big)_{2i}.
    \end{aligned}
    \label{eq:allocation_inverse}
\end{equation}
The relationship between the servo angular velocity and wrench change rate shown in Eq.~\eqref{eq:allocation_inverse} reveals a critical insight into the system dynamics. When the wrench changes rapidly, it directly impacts the required servo angular velocity. Since servos typically have slower dynamics compared to the BLDC motors, excessively rapid wrench changes can cause the servos to lag behind commanded angles. This lag creates a mismatch between the commanded and actual thrust vectors, potentially leading to oscillations and instability in the system. Therefore, the wrench change rate must be carefully constrained in the controller to remain within the physical limitations of the servo dynamics, ensuring smooth and stable performance of the Tilt-Ropter system.

%%%%%%%%%%%%%%%%%%%%%%%%%%%%%%%%%%%%%%%%%%%%%%%%%%%%%%%%%%%%%%%%%%%%%%%%%%%%%%%%

\section{Control}
\label{Control Design}

\begin{figure}[tbp]
    \centering
    \includegraphics[width=1.0 \linewidth]{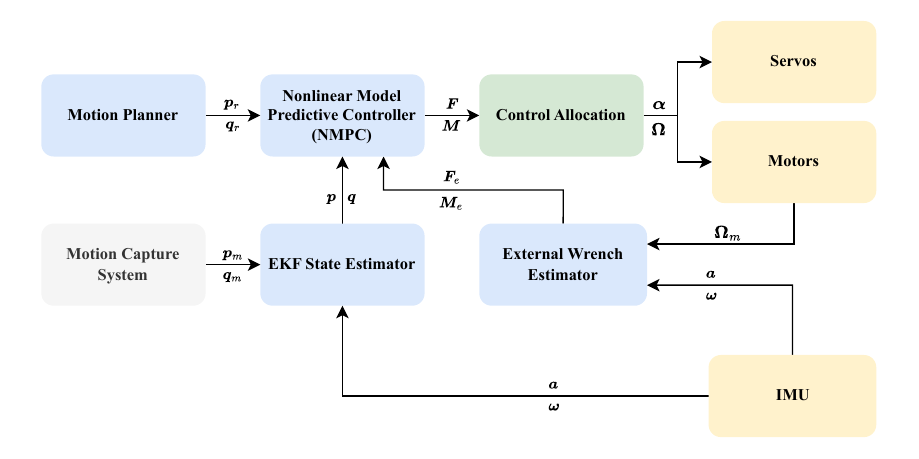}
    \caption{Control architecture of Tilt-Ropter. $\vec{p}_m$ and $\vec{q}_m$ are the measured position and quaternion of the robot from motion capture system. $\vec{\Omega}_m$ is the measured rotor speeds from ESC.}
    \vspace{-12pt}
    \label{fig:control_diagram}
\end{figure}

The overall control architecture is illustrated in Fig.~\ref{fig:control_diagram}. 
Reference trajectories generated by the motion planner are tracked using a nonlinear model predictive controller. NMPC is chosen to address the nonlinear rigid-body dynamics, actuation redundancy, and mode-dependent non-holonomic constraints within a unified optimization framework.

Based on the control allocation formulation in Sec.~\ref{sec:control allocation}, the NMPC generates wrench commands $\vec{W}$ instead of directly optimizing rotor speeds and servo angles. Although actuator-level NMPC could fully exploit the platform's actuation redundancy, its increased input dimensionality and computational burden limit real-time feasibility.
The wrench-based formulation achieves a favorable trade-off between computational efficiency and actuator feasibility, while generally improving power efficiency and tracking accuracy except under highly aggressive maneuvers \cite{brunner2022mpc}.

The optimized wrench is mapped to actuator commands through control allocation and executed by the motors and servos. 
An Extended Kalman Filter (EKF) provides state estimation by fusing inertial and motion-capture measurements, while ground-contact effects are estimated by an actuator-aware external wrench estimator and incorporated into the control framework.
Together, these components ensure accurate trajectory tracking and robust performance across locomotion modes.

\subsection{Nonlinear Model Predictive Control}

The control problem is formulated as a finite-time optimal control problem (OCP), where the objective is to minimize the tracking error of the robot state and input to the reference trajectory while satisfying the system dynamics and constraints. Additionally, NMPC enables smooth interaction between the robot and the ground by incorporating contact wrenches into the dynamic model and formulating ground-specific constraints within the control problem. This approach provides a unified control architecture for both aerial and terrestrial locomotion, facilitating seamless transitions between modes.

\subsubsection{States and Inputs}
Firstly, we choose the system state $\vec{x}$ and control inputs $\vec{u}$ as
\begin{subequations}
    \begin{align}
    \vec{x}&=\matrix{\vec{F}\transpose & \vec{M}\transpose & \vec{p}\transpose & \vec{v}\transpose & \vec{q}\transpose & \vec{\omega}\transpose}\transpose\in\nR{19}, \\
    \vec{u}&=\matrix{\dot{\vec{F}}\transpose & \dot{\vec{M}}\transpose}\transpose\in\nR{6}.
    \end{align}
\end{subequations}
Note that the NMPC does not directly optimize the wrench command $\vec{W}$, but rather its derivative $\dot{\vec{W}}$ as the control input. This choice implicitly incorporates the actuator dynamics, particularly the slower servo responses, into the optimization process.

\subsubsection{Problem Formulation}
Accordingly, the NMPC optimization problem is formulated as
\begin{subequations}
    \begin{align}
        \begin{split}
            \vec{u}_{\text{NMPC}} = & \argmin_{\vec{u}}  \sum_{k=0}^{N-1} \left( \tilde{\vec{x}}_k^\top \mat{Q} \tilde{\vec{x}}_k + \vec{u}_k^\top \mat{R} \vec{u}_k \right) \\
            & \qquad \qquad + \tilde{\vec{x}}_N^\top \mat{Q}_N \tilde{\vec{x}}_N 
        \end{split}\label{eq:mpc_optimization}\\
        \begin{split}
            \text{subject to} \quad
            & \vec{x}_0 = \vec{x}_{est}, \\
            & \vec{x}_{k+1} = \vec{f}(\vec{x}_k, \vec{u}_k), \\
            & \vec{u}_k \in [-\vec{u}_{max}, \vec{u}_{max}], \\
            & \vec{W}_k \in [\vec{W}_{min}, \vec{W}_{max}], \\
            & \delta \cdot v_{\scriptscriptstyle B, y} = 0, \\
            & \delta \cdot \omega_{x} = 0, \\
            & \delta \cdot (p_{z} - r) = 0, \\
        \end{split} \label{eq:mpc_subject}
    \end{align} \label{eq:mpc}
\end{subequations}
\par\noindent where $N$ is the length of prediction horizon. $\mat{Q}$ and $\mat{Q}_N$ are the stage and terminal state cost matrices, respectively. $\mat{R}$ is the cost matrix for the control input. $\vec{x}_{est}$ is the current estimated state of the robot. The system dynamics are described by $\vec{f}$, the discrete-time formulation of the continuous dynamics given in \eqref{eq:rb1}--\eqref{eq:rb4}. The input bounds $\vec{u}_{max}$ restrict the rate of change of the commanded wrench. These bounds also implicitly enforce the servo rate limitation derived in Sec.~\ref{sec:control allocation}, ensuring that the generated actuator commands remain dynamically feasible. $\vec{W}_{min}$ and $\vec{W}_{max}$ are the constraints on the wrench command, ensuring robot safety and stability.

The non-holonomic constraints on the ground are formulated using an indicator function $\delta$, defined as:
\begin{equation}
    \delta = \left\{
    \begin{aligned}
        &1, && \text{if } p_{r,z} = r \text{ and } p_z = r \\[4pt]
        &0, && \text{otherwise}
    \end{aligned}
    \right.
\end{equation}
where $p_{r,z}$ is the reference height from motion planner, $p_z$ is the current height of the robot, and $r$ is the radius of the wheel. The indicator function $\delta$ is used to activate or deactivate the non-holonomic constraints based on the robot's state. Only when the robot is on the ground ($p_z = r$) and the reference trajectory is a ground trajectory ($p_{r,z} = r$), the non-holonomic constraints are activated. In Eq. \eqref{eq:mpc_subject}, the last three constraints impose the non-holonomic conditions when the robot is on the ground. Specifically, the first constraint ensures that the robot does not move in the $y$-direction. Note that the velocity $v_{\scriptscriptstyle B, y}$ is the $y$-component of the velocity in the body frame $\frame{B}$. The second constraint prevents rotation about the $x$-axis. The third constraint ensures that the robot maintains contact with the ground by keeping the height equal to the wheel radius. These constraints are crucial for maintaining stability and preventing undesired motion during ground locomotion.

The state tracking error $\tilde{\vec{x}}$ at time step $k$ is defined as
\begin{equation}
    \tilde{\vec{x}}_k =
    \begin{bmatrix}
        \vec{F}_k \\ \vec{M}_k \\
        \vec{p}_k - \vec{p}_{r,k} \\ \vec{v}_k - \vec{v}_{r,k} \\
        \tilde{\vec{x}}_{\vec{q},k} \\
        \vec{\omega}_k - \vec{\omega}_{r,k}
    \end{bmatrix},
\end{equation}
where $\vec{p}_{r,k}$, $\vec{v}_{r,k}$, $\vec{q}_{r,k}$, and $\vec{\omega}_{r,k}$ are the reference position, velocity, quaternion, and angular velocity obtained from the trajectory planner. Since unit quaternions lie on a nonlinear manifold, orientation error is defined as $\tilde{\vec{x}}_{\vec{q},k} = \vec{q}_k \otimes \vec{q}_{r,k}^{-1}$, representing the minimal rotation that aligns the current orientation with the reference.

%%%%%%%%%%%%%%%%%%%%%%%%%%%%%%%%%%%%%%%%%%%%%%%%%%%%%%%%%%%%%%%%%%%%%%%%%%%%%%%%

\section{External Wrench Estimation}
\label{External Wrench Estimation}

External wrench estimation is essential for Tilt-Ropter, particularly during ground locomotion where wheel-ground interactions introduce complex forces that are difficult to model. Instead of constructing a detailed contact model, these effects are treated as external wrenches acting on the robot. The estimator is based on the framework in \cite{tomic2017external}, which requires only acceleration and angular velocity measurements. Building upon the original formulation, the applied actuator wrench is reconstructed using measured rotor speeds and estimated servo angles. This extension is specifically tailored to the tilt-rotor MAVs with servo mechanisms, allowing actuator dynamics to be incorporated into the estimation process.

The external wrench estimation can be expressed as
\begin{equation}
\begin{aligned}
    \hat{\vec{F}}_e &= \mat{K}_f \int \left(m \vec{a} - \vec{F} - \hat{\vec{F}}_e \right) \mathrm{d}t, \\
    \hat{\vec{M}}_e &= \mat{K}_m \left( \bm{J} \vec{\omega} - \int \left( \vec{M} + (\bm{J} \vec{\omega}) \times \vec{\omega} + \hat{\vec{M}}_e \right) \mathrm{d}t \right),
\end{aligned}
\label{eq: wrench_estimation}
\end{equation}
where $\vec{a} = \vec{q}^{-1} \odot (\dot{\vec{v}} - \vec{g})$ is the measured acceleration of the robot in the body frame. $\hat{\vec{F}}_e$ and $\hat{\vec{M}}_e$ are the estimated external force and torque acting on the robot, respectively. $\mat{K}_f$ and $\mat{K}_m$ are the estimation gains.

This estimator reconstructs the external wrench by comparing the robot motion inferred from inertial measurements with the wrench generated by the actuators. Specifically, the force component is estimated from the residual between the measured body-frame acceleration and the actuator-generated force, while the torque component follows a momentum-based formulation to avoid direct numerical differentiation of the angular velocity. This formulation helps reduce the influence of high-frequency measurement noise while maintaining sensitivity to the main interaction effects caused by wheel-ground contact and other unmodeled disturbances.

To obtain the actuator-generated force $\vec{F}$ and torque $\vec{M}$, relying solely on the commanded wrench from the NMPC is insufficient, as it does not account for actuator dynamics. Instead, the applied wrench is reconstructed from motor and servo states. The rotor speeds are measured from the ESC, whereas the servo angles are estimated using a first-order response model.

\begin{figure}[htbp]
    \centering
    \vspace{6pt}
    \includegraphics[width=0.95 \linewidth]{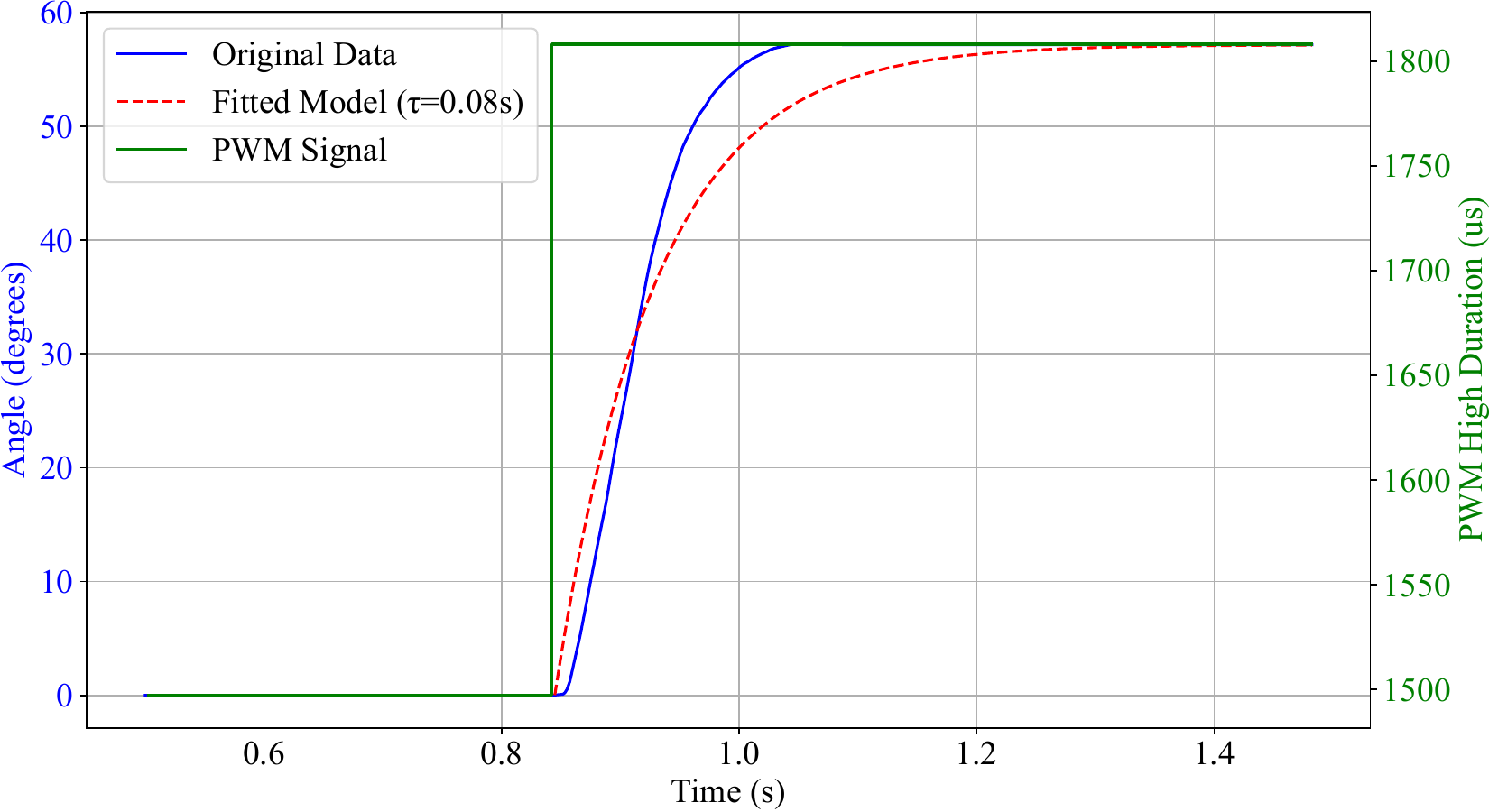}
    \caption{System identification of the servo with attached motor and propeller. The green line represents the PWM input signal to the servo, the blue line shows the original angle data collected by the rotary encoder, and the red line depicts the fitted response based on the identified first-order model.}
    \vspace{-6pt}
    \label{fig:servo_id}
\end{figure}

Since most low-cost servo motors do not provide angle feedback, the actual servo angles cannot be directly measured. A first-order dynamic model is therefore adopted to estimate the servo response from the commanded tilt angle $\alpha$. The estimated servo angle $\hat{\alpha}$ is modeled as
\begin{equation}
    \dot{\hat{\alpha}} = \frac{1}{\tau}(\alpha - \hat{\alpha}),
\end{equation}
where $\alpha$ denotes the commanded servo angle from the control allocation, $\hat{\alpha}$ denotes the estimated servo angle, and $\tau$ is the servo time constant. The parameter $\tau$ is identified via step response experiments, as shown in Fig.~\ref{fig:servo_id}, where the recorded angular response is fitted to a first-order model using least squares estimation.

With the servo dynamics characterized as a first-order system and the time constant identified, the actual servo angles can be estimated by simulating the servo response to the commanded inputs. These angles, together with the measured rotor speeds, are substituted into Eq.~\eqref{eq: allocation} to compute the total force $\vec{F}$ and torque $\vec{M}$ generated by the actuators. The external wrench, comprising $\hat{\vec{F}}_e$ and $\hat{\vec{M}}_e$, is then obtained by substituting the computed $\vec{F}$ and $\vec{M}$ into Eq.~\eqref{eq: wrench_estimation}. This approach improves estimation accuracy by accounting for actuator dynamics and the absence of direct servo angle measurements.

%%%%%%%%%%%%%%%%%%%%%%%%%%%%%%%%%%%%%%%%%%%%%%%%%%%%%%%%%%%%%%%%%%%%%%%%%%%%%%%%

\section{Experiments and Results}
\label{Tilt-Ropter Experimental Results}
\subsection{Simulation}
To validate the proposed robot configuration and control algorithms, we developed a comprehensive simulation platform for Tilt-Ropter using the Gazebo simulator. We carefully measured and calculated key physical parameters such as mass, CoM, and inertia, ensuring a minimal gap between the simulation and real-world performance. Different from the traditional quadrotor simulation, our simulation incorporates both servo dynamics and the motion of passive wheels. This approach allows us to accurately simulate the tilt-rotor mechanisms and the generation of vector thrust. Additionally, the interaction between the wheels and the ground, including contact forces, is managed by the Open Dynamics Engine (ODE) within Gazebo, providing a realistic representation of on-ground dynamics.

\begin{figure}[htbp]
    \centering
    \includegraphics[width=1.0 \linewidth]{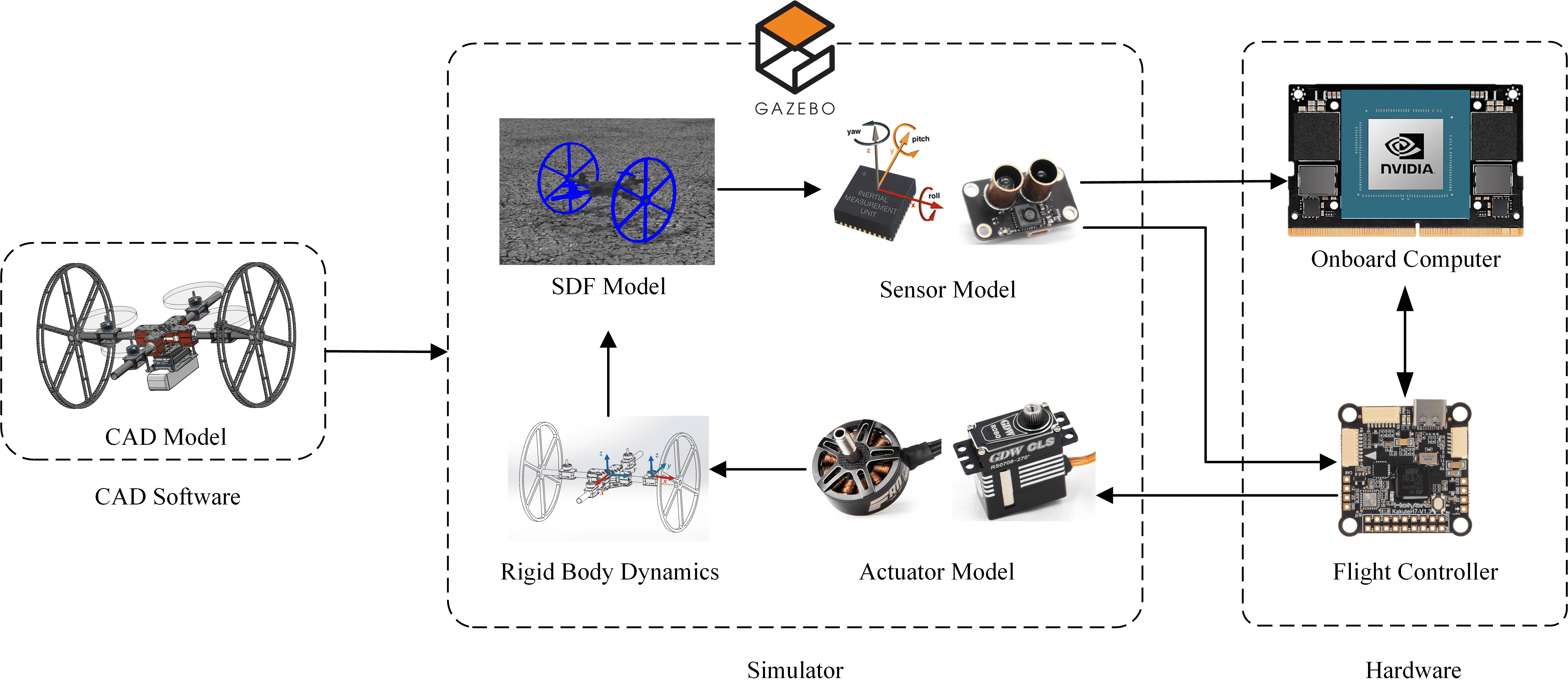}
    \caption{The structure of Hardware-in-the-Loop simulation.}
    \vspace{-6pt}
    \label{fig:hitl}
\end{figure}

Besides the common Software-in-the-Loop (SITL) simulation, we also developed a Hardware-in-the-Loop (HITL) simulation to test the algorithms with hardware on the real robot. The structure of HITL simulation is shown in Fig.~\ref{fig:hitl}. In our HITL simulation, the real robot is integrated with the simulation environment via the ROS framework. The Gazebo simulator generates virtual sensor data, which is utilized by the state estimation module in the physical onboard computer. Simultaneously, this onboard computer processes high-level control commands using NMPC algorithms. The control commands are then forwarded to the physical low-level flight controller, which generates actuator commands for the simulated actuators. The state of the simulated robot is dynamically updated based on these commands, guided by the rigid body dynamics model implemented within the simulation. This integration ensures a continuous and accurate exchange of data and control between the real and simulated components, significantly enhancing both the fidelity and effectiveness of the simulation. A demonstration of the simulation results, including air-ground transitions and trajectory tracking, is provided in the supplementary video.

\subsection{Real-World Experiments}

We conducted a series of real-world experiments to validate the performance of the Tilt-Ropter. The robot was tested in various scenarios, including aerial flights, ground locomotion, and mode transitions. The results demonstrated the effectiveness of the proposed model-based control algorithms and the versatility of the Tilt-Ropter design. The parameters for the NMPC in these experiments are shown in Table~\ref{tab:real_mpc_parameters}.

\begin{table}[htbp]
    \small
    \centering
    \caption{NMPC Parameters for Real-World Experiments}
    \begin{tabular}{l|l}
    \hline
    Parameter & Value \\ \hline
    $\dot{\vec{F}}_{max}$    & (2, 2, 2)\, $\mathrm{N\cdot s^{-1}}$ \\
    $\dot{\vec{M}}_{max}$   & (2, 2, 2)\, $\mathrm{N\cdot m\cdot s^{-1}}$ \\
    $\vec{F}_{min}$       & (-0.2, -0.2, 0.0)\, $\mathrm{N}$ \\
    $\vec{F}_{max}$       & (0.2, 0.2, 20)\, $\mathrm{N}$ \\
    $\vec{M}_{min}$      & (-20, -20, -20)\, $\mathrm{N\cdot m}$ \\
    $\vec{M}_{max}$      & (20, 20, 20)\, $\mathrm{N\cdot m}$ \\
    $\omega_{max}$         & (2.0, 2.0, 1.5)\, $\mathrm{rad\cdot s^{-1}}$ \\
    $N$               & 20 \\
    $\Delta t$        & 0.1\, $\mathrm{s}$ \\
    \hline
    \end{tabular}
    \label{tab:real_mpc_parameters}
\end{table}

\subsubsection{Aerial Flight}

\begin{figure}[htbp]
    \vspace{6pt}
    \centering
    \includegraphics[width=0.86 \linewidth]{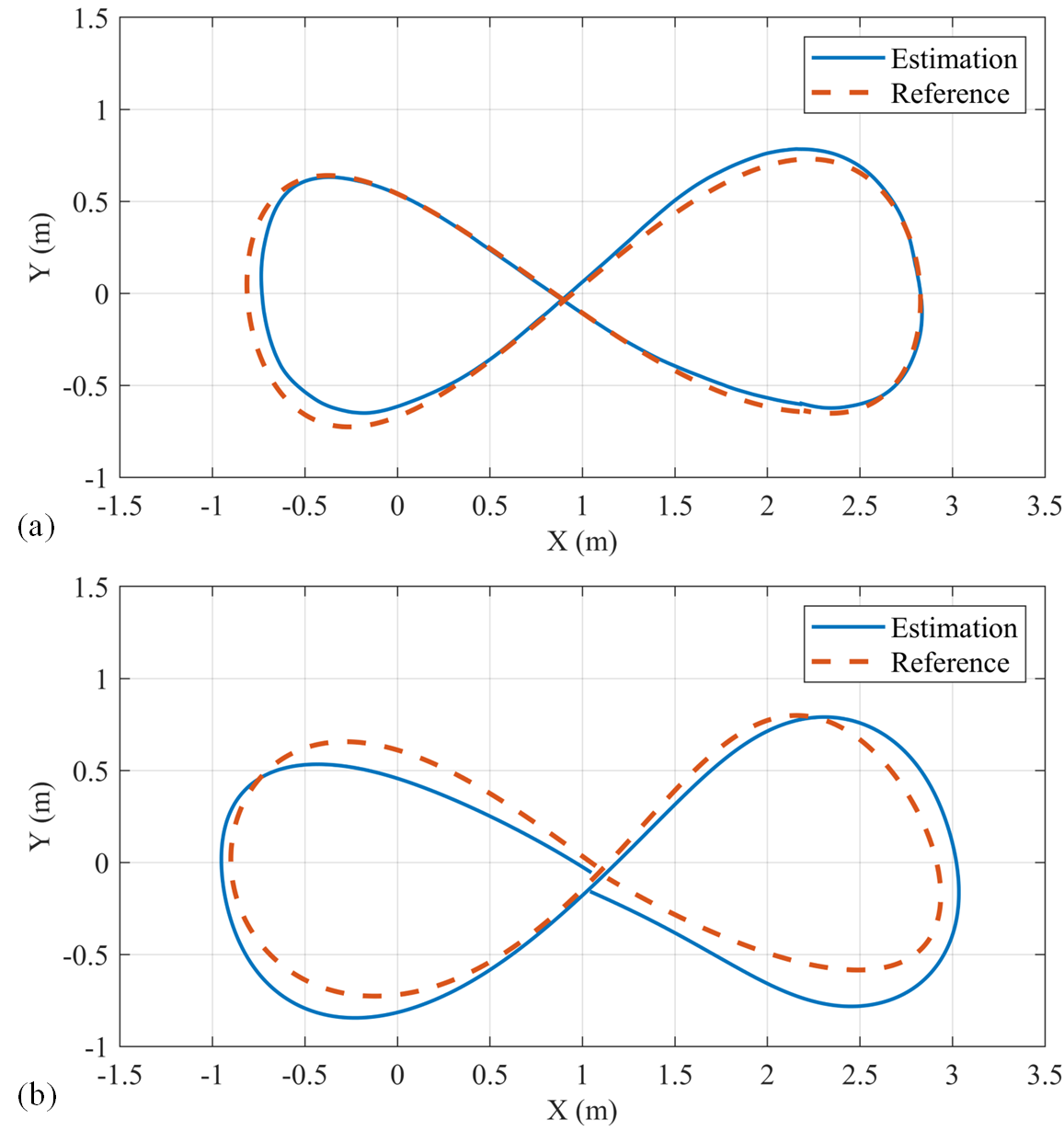}
    \caption{Trajectories of the Tilt-Ropter in aerial and ground modes. (a) Trajectory of Tilt-Ropter in aerial flight. (b) Trajectory of Tilt-Ropter in ground locomotion.}
    % \vspace{-6pt}
    \label{fig:exp_air_or_gnd}
\end{figure}

To validate the aerial flight performance of the Tilt-Ropter, a trajectory tracking experiment was conducted. The robot was commanded to follow the trajectory as shown in Fig.~\ref{fig:exp_air_or_gnd}(a). 

The trajectory is a figure-eight shape, which includes both horizontal and vertical movements. The maximum velocity and acceleration of the robot are set to 1.5 m/s and 1.5 m/s$^2$ respectively. The robot successfully tracked the trajectory with Root Mean Square Error (RMSE) of 0.052 m, demonstrating the effectiveness of the NMPC-based control algorithms.

\subsubsection{Ground Locomotion}

Similar to the aerial flight experiment, we conducted a ground locomotion experiment to evaluate the performance of the Tilt-Ropter in terrestrial mode. The only difference from the aerial-flight experiment is that the robot is commanded to move on the ground. The robot successfully tracked the trajectory with RMSE of 0.145 m, as shown in Fig.~\ref{fig:exp_air_or_gnd}(b), demonstrating the effectiveness of the proposed control algorithms in ground locomotion scenarios.

\subsubsection{Air-Ground Mode Transition}

\begin{figure}[htbp]
    \vspace{-6pt}
    \centering
    \includegraphics[width=0.9 \linewidth]{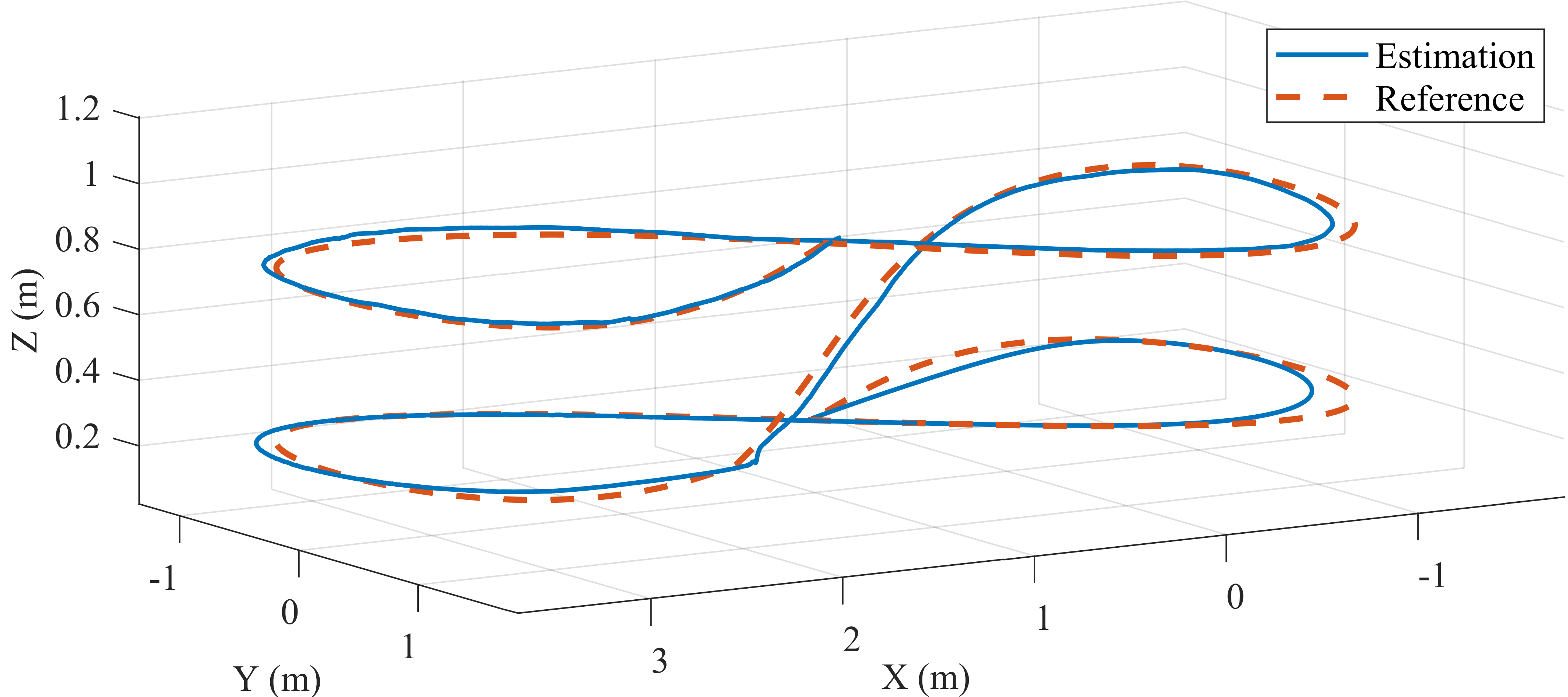}
    \caption{Hybrid trajectory of Tilt-Ropter showcasing seamless air-ground mode transitions.}
    \label{fig:exp_air_gnd}
    \vspace{-12pt}
\end{figure}

To evaluate the air-ground mode transition performance of the Tilt-Ropter, we conducted a trajectory tracking experiment encompassing both aerial flight and ground locomotion phases. The robot starts from an arbitrary position in the air, follows a figure-eight trajectory with air locomotion mode, then transitions smoothly to the ground, continuing to track a similar figure-eight trajectory on the ground. The trajectory is shown in Fig.~\ref{fig:exp_air_gnd}. The robot successfully completed the trajectory and performed a smooth mode transition between aerial flight and ground locomotion with RMSE of 0.125 m, demonstrating the versatility of the NMPC controller and robustness of the Tilt-Ropter design.

\begin{figure}[htbp]
    \vspace{6pt}
    \centering
    \includegraphics[width=1.0 \linewidth]{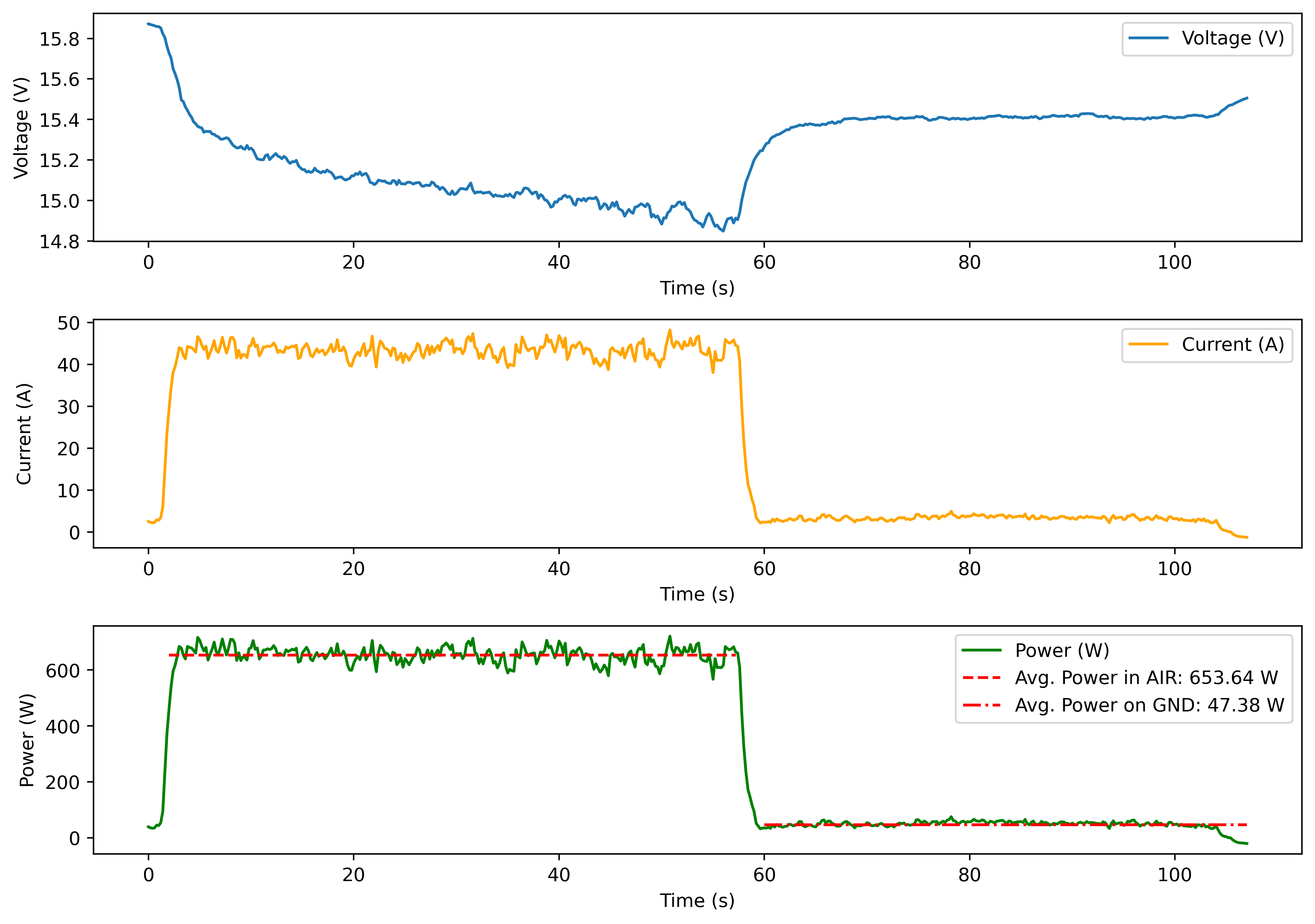}
    \caption{Comparison of power consumption between aerial flight and ground locomotion.}
    \label{fig:power}
    \vspace{-12pt}
\end{figure}

\subsubsection{Energy Efficiency}

To analyze the energy efficiency of the Tilt-Ropter design, we compared the power consumption of the robot during aerial flight and ground locomotion scenarios. The robot is given a hybrid trajectory that includes both aerial and ground phases. The power consumption is calculated by multiplying the voltage and current of the battery measured during the experiments. The voltage of the battery is measured by the Analog-to-Digital Converter (ADC) inside the flight controller, and the current is measured by the current sensor within the ESC. The results are shown in Fig.~\ref{fig:power}. We calculated the average power consumption during the aerial flight and ground locomotion phases, which are 653.64 W and 47.38 W, respectively. The results indicate that the power consumption of the robot in ground locomotion is only 7.2\% of that in aerial flight. This significant reduction in power consumption during ground locomotion highlights the energy efficiency of the Tilt-Ropter design, which makes it an ideal platform for long-endurance missions by leveraging its multi-modal capabilities.

%%%%%%%%%%%%%%%%%%%%%%%%%%%%%%%%%%%%%%%%%%%%%%%%%%%%%%%%%%%%%%%%%%%%%%%%%%%%%%%%

\section{Conclusion}
\label{Conclusion}

In this work, we presented the design, modeling, and control of Tilt-Ropter, a fully actuated hybrid aerial-terrestrial vehicle. By integrating tilt rotors with passive wheels, the platform achieves energy-efficient ground locomotion while preserving aerial agility and decoupled force-torque control. A unified model-based control framework leverages the fully actuated design to achieve trajectory tracking and seamless transitions across aerial and terrestrial modes, while a dedicated control allocation ensures actuator feasibility by implicitly accounting for servo rate limitations. To address complex wheel-ground interactions, an actuator-aware external wrench estimation method was incorporated into the control architecture. The proposed system was validated through Hardware-in-the-Loop simulations and real-world experiments, demonstrating reliable multi-modal locomotion, smooth mode transitions, and significant energy savings during ground operation.


%%%%%%%%%%%%%%%%%%%%%%%%%%%%%%%%%%%%%%%%%%%%%%%%%%%%%%%%%%%%%%%%%%%%%%%%%%%%%%%%

%%%%%%%%%%%%%%%%%%%%%%%%%%%%%%%%%%%%%%%%%%%%%%%%%%%%%%%%%%%%%%%%%%%%%%%%%%%%%%%%

%%%%%%%%%%%%%%%%%%%%%%%%%%%%%%%%%%%%%%%%%%%%%%%%%%%%%%%%%%%%%%%%%%%%%%%%%%%%%%%%

%%%%%%%%%%%%%%%%%%%%%%%%%%%%%%%%%%%%%%%%%%%%%%%%%%%%%%%%%%%%%%%%%%%%%%%%%%%%%%%%

\bibliographystyle{conf/IEEEtran}
\balance
\bibliography{reference}

\end{document}